\documentclass[letterpaper]{article} %
\usepackage{aaai25}  %
\usepackage{times}  %
\usepackage{helvet}  %
\usepackage{courier}  %
\usepackage[hyphens]{url}  %
\usepackage{graphicx} %
\usepackage{natbib}  %
\usepackage{caption} %
\usepackage{algorithm}
\usepackage{algorithmic}

\usepackage{graphicx}
\usepackage{subcaption}
\usepackage{url}
\usepackage{newfloat}
\usepackage{listings}
\DeclareCaptionStyle{ruled}{labelfont=normalfont,labelsep=colon,strut=off} %
\floatstyle{ruled}
\newfloat{listing}{tb}{lst}{}
\floatname{listing}{Listing}
\title{An Automated Explainable Educational Assessment System Built on LLMs}
\author{
    Jiazheng Li\textsuperscript{\rm 1}\equalcontrib, Artem Bobrov\textsuperscript{\rm 1}\equalcontrib, David West\textsuperscript{\rm 2}, Cesare Aloisi\textsuperscript{\rm 2}, Yulan He\textsuperscript{\rm 1,3}\\
}
\affiliations{
    \textsuperscript{\rm 1}King's College London,\\
    \textsuperscript{\rm 2}AQA,\\
    \textsuperscript{\rm 3}The Alan Turing Institute\\
    \{jiazheng.li, artem.bobrov, yulan.he\}@kcl.ac.uk,~~~\{dwest, caloisi\}@aqa.org.uk
}

\begin{document}
\maketitle

\begin{abstract}
In this demo, we present AERA Chat, an automated and explainable educational assessment system designed for interactive and visual evaluations of student responses. This system leverages large language models (LLMs) to generate automated marking and rationale explanations, addressing the challenge of limited explainability in automated educational assessment and the high costs associated with annotation. Our system allows users to input questions and student answers, providing educators and researchers with insights into assessment accuracy and the quality of LLM-assessed rationales. Additionally, it offers advanced visualization and robust evaluation tools, enhancing the usability for educational assessment and facilitating efficient rationale verification.
\end{abstract}

\section{Introduction}
Automated Student Answer Scoring (ASAS) systems are vital educational NLP applications that streamline the manual grading process, offering a swift and consistent evaluation of student performance \cite{grading_classification, helen-aes-2016, yue-aes-2017}. Traditional ASAS systems typically utilize text classifiers built on pre-trained language models \cite{bert_classifer_aes, xie-etal-2022-automated}. These models process inputs such as question, key answer elements (e.g., ``\textit{student should specify the materials to be tested in their response}''), marking rubric (e.g., ``\textit{2 points for accurately describing two additional pieces of information}''), and student responses to generate marks. However, concerns about the lack of transparency of these systems have raised questions about their reliability in real-world assessments. Various approaches have been developed to address this issue by interpreting the model marking processes. These include feature analysis \cite{tornqvist-etal-2023-exasag, bert_feature, pmlr-v216-li23d} and visualizations of internal mechanisms, such as weights and attention \cite{helen-aes-2016, yang-etal-2020-enhancing}. However, these interpretations often require significant NLP expertise, posing challenges for educators without technical backgrounds.

The recent development of large language models (LLMs) has introduced a new approach that leverages in-context learning and reasoning capabilities \cite{few_shot, zero_shot,zhou-etal-2024-mystery} to generate natural language rationales that justify model decisions \cite{oana-esnli, few-shot-rationalization, rationalization_survey}. This approach improves explainability, making it more accessible to both educators and students. However, acquiring rationale annotations is costly. Most ground truth labels used for training rationale generation models rely on noisy rationales generated by LLMs, often without human verification.

To \textbf{\emph{facilitate easy access to the latest LLM-based assessment rationales generation models}} and \textbf{\emph{simplify the rationale evaluation and annotation process}}, we introduce an interactive platform for \textbf{A}utomated \textbf{E}xplainable student \textbf{R}esponse \textbf{A}ssessment, called \textbf{AERA Chat}. Our platform features an interactive user interface, leveraging multiple LLMs as the backend for automated assessment and rationale generation. AERA Chat is designed to advance the practical use of explainable assessment methods in educational scenarios and support research by providing a unified platform for benchmarking. To the best of our knowledge, AERA Chat is the first interactive platform designed to utilize LLMs for explainable student answer scoring. Our demo video can be found at \url{https://youtu.be/qUSjz-sxlBc}.

\vspace{-2mm}
\section{Overview of AERA Chat Functionality}

\begin{figure*}[ht]
    \centering
    \begin{subfigure}[b]{0.350\linewidth}
        \includegraphics[width=\linewidth]{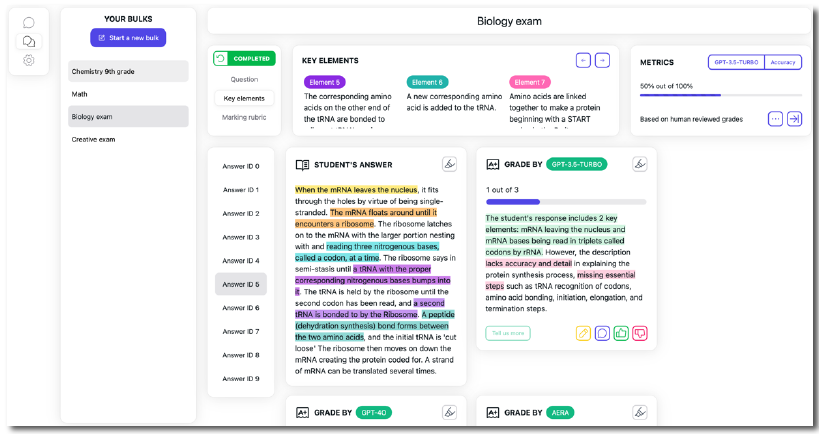}
        \caption{Overview of Bulk Marking Interface.}
        \label{fig:chat_interface}
    \end{subfigure}
    \begin{subfigure}[b]{0.645\linewidth}
        \includegraphics[width=\linewidth]{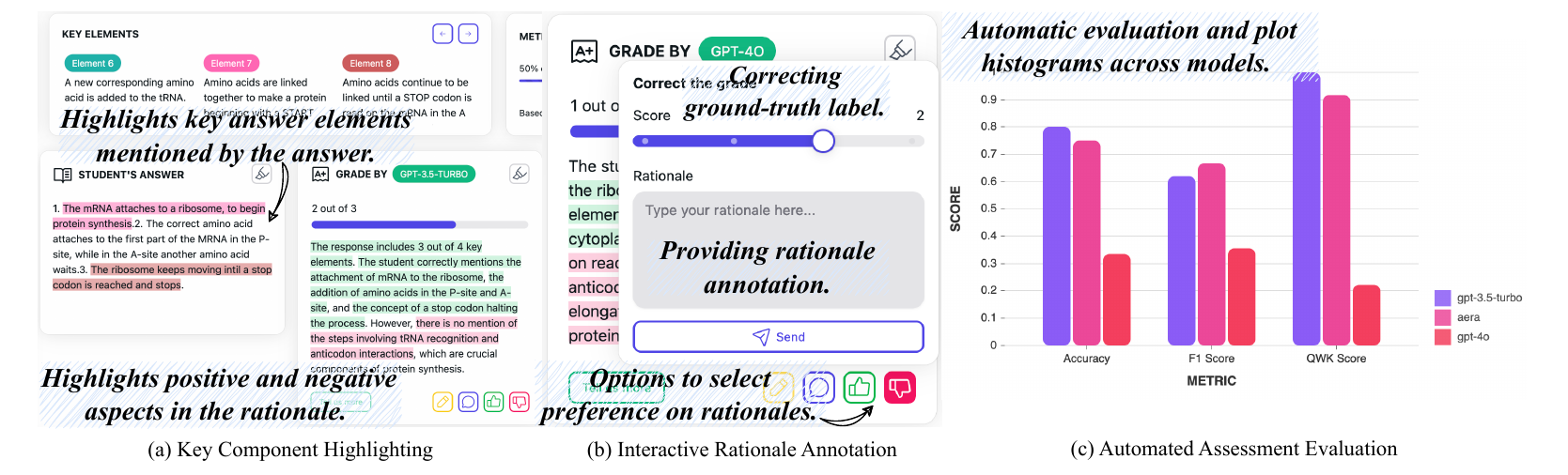}
        \caption{Key Functionalities Provided in the Bulk Marking Interface.}
        \label{fig:bulk_marking}
    \end{subfigure}
    \caption{The AERA Chat Interface.}
    \vspace{-4mm}
\end{figure*}

\subsection{Bulk Marking Interface}
As shown in Fig.\ref{fig:chat_interface}, the bulk marking interface is our main interface to provide automated assessment and integrate with our innovative tools. Our interface has been implemented and equipped with the following key functionalities:

\paragraph{Bulk Question Initialization} Users can set up the question, key answer elements, and a point-based marking rubric in the system. Our system automatically compiles the question information into a template prompt, which is then provided to LLMs for assessment. Users can upload a batch of student answers they want to assess in a file, which may or may not include the ground-truth marks. 

\paragraph{Automated Scoring and Rationale Generation}
Once the question information and student answers have been uploaded to the system, our platform can automatically assess student answers and generate rationales with the user-selected LLMs. In this system demonstration, we used two OpenAI models: \texttt{GPT-3.5-turbo} and \texttt{GPT-4o}, as well as our developed rationale generation model -- the AERA model \cite{li-etal-2023-distilling}. Our system can easily extend to include other models for student answer assessment and rationale generation. Our database system will automatically monitor the assessment status. As shown in Fig.\ref{fig:chat_interface}, our platform will display rationales and marks assessed by each LLM as a card view once the assessment is completed. %

\paragraph{Explainable Highlighting}
Verifying the faithfulness of the assessment rationales for student answers is a complex task that requires a detailed understanding of context. Our platform aims to enhance user experience by providing clear, high-contrast visual cues within student answers and assessment rationales. As shown in the leftmost picture in Fig.\ref{fig:bulk_marking}, users can choose to visualize the key answer elements mentioned in the student answers or visualize the positive aspects (reasons for awarding points) and negative aspects (reasons for deducting points) within rationales. %
This functionality is powered by GPT-4o, which performs word-level tagging and generates context highlights in a JSON format, ensuring effective and efficient context visualization  \cite{overprompt}.

\paragraph{Annotation Toolkit} To enable easy access for rationale evaluation and annotation, we have implemented three functionalities, as demonstrated in the middle picture in Fig. \ref{fig:bulk_marking}. \textbf{(1)} \underline{Ground-truth Label Correction}: As noted in \cite{li-etal-2023-distilling}, some ground-truth labels provided by publicly available ASAS dataset \cite{asap-aes} could be wrong. Therefore, we offer users an option to correct wrong ground-truth labels. \textbf{(2)} \underline{Human Preference Annotation}: Evaluating the quality of rationales often models the qualitative evaluation task as binary preferences \cite{2024_thought_tree, Li2024AERACA}, where the factually correct and more detailed rationale is preferred. Consequently, our platform includes options for users to select ``preferred'' or ``not preferred'' rationales. These selections are automatically recorded in the system's database, allowing researchers to construct preference data that can be used to leverage the reinforcement learning from human feedback techniques \cite{dpo,lu-etal-2024-eliminating,rlhf} for training LLMs for rationale generation. \textbf{(3)} \underline{Rationale Annotation}: If none of the candidate rationales generated by LLMs is correct, our platform allows users to submit their assessment rationales as annotated data for future supervised fine-tuning LLMs.

\paragraph{Assessment Performance Evaluation}
As shown in the rightmost picture of Fig.\ref{fig:bulk_marking}, our platform can automatically evaluate LLMs' assessment performance if the user uploads their pre-assessed marks. This evaluation is visualized in a histogram displaying Accuracy, macro F1 Scores, and QWK (Quadratic Weighted Kappa) scores. 

\subsection{Chat Interface} \label{sec:chat_marking}
To leverage the advanced chat capabilities of LLMs and their potential to support explainable student answer scoring, our platform integrates a chat interface. This feature allows users to bring question details and rationales from the bulk marking system into interactive discussion with LLMs. Educators can use this functionality to request more detailed explanations for unclear assessment rationales, while researchers can use it to analyze incorrect assessment rationales and regenerate assessment decisions for further refinement. %

\vspace{-2mm}
\section{System Implementation Details}
AERA Chat is designed using a microservices architecture that integrates multiple LLMs through a unified web interface. We use Docker to modularize services. \textbf{Frontend}:
The frontend provides a responsive web interface tailored for educators and researchers to engage with the system. It was developed using the Remix framework, which is built on React.
\textbf{Backend}: The backend layer serves as the backbone of the AERA Chat platform, integrating all other layers. It operates based on a REST API, handling requests over HTTP connections without maintaining the state of each connection. This layer is implemented using the Flask framework. \textbf{LLM Services}: Users of our system can opt to employ publicly available API-based LLMs, or they may choose to develop and utilize privately trained, customized LLMs to enhance question-specific performance or ensure better data privacy. We developed a local API module allowing users to deploy privately trained models using the HuggingFace package as their private API service. \textbf{Database}: We utilize a PostgreSQL relational database to manage various data, including user profiles, assessment records (both for active tracking and background processing), and chat histories. 

\vspace{-2mm}
\section{Conclusion and Future Work}
In conclusion, we have developed an interactive platform, AERA Chat, for explainable student answer assessment via rationale generation. Our platform supports public and private LLM assessment, uses highlighting techniques for better visualization, and includes annotation functions to help educators and researchers generate more accurate data for developing trustworthy automated assessment LLMs. With a potential for wider applications to enable trustworthy automated student answer scoring, AERA Chat paves the way for future advancements of LLMs in educational settings.

\section{Acknowledgments}
This work was supported in part by the UK Engineering and Physical Sciences Research Council through KCL's Impact Acceleration Account (grant no. EP/X525571/1) and a Turing AI Fellowship (grant no. EP/V020579/1, EP/V020579/2). JL is funded by a PhD scholarship provided by AQA.   

\bibliography{aaai25}

\end{document}